\documentclass[11pt,a4paper]{article}
\pdfoutput=1
\usepackage[nohyperref]{emnlp2018}
\usepackage{hyperref}
\usepackage[pdftex]{xcolor}
\usepackage{graphicx}
\usepackage{times}
\usepackage{latexsym}
\usepackage{amsmath,amssymb}
\usepackage{booktabs}
\usepackage{stmaryrd}
\usepackage{url}
\usepackage{lipsum}
\usepackage{xargs} 
\usepackage{tikz} 
\usetikzlibrary{arrows,automata,positioning}
\usetikzlibrary{shadows}

\definecolor{darkblue}{rgb}{0, 0, 0.5}
\hypersetup{colorlinks=true,citecolor=darkblue, linkcolor=darkblue, urlcolor=darkblue}

\aclfinalcopy
\def\aclpaperid{16} 

\input{macros.inp}
\title{Jump to better conclusions: SCAN both left and right}

\author{Jasmijn Bastings\textsuperscript{1} \quad Marco Baroni\textsuperscript{2} \quad Jason Weston\textsuperscript{2} \quad Kyunghyun Cho\textsuperscript{2,3,4} \quad Douwe Kiela\textsuperscript{2}\\
\textsuperscript{1}ILLC, University of Amsterdam\\ \textsuperscript{2}Facebook AI Research\\
\textsuperscript{3}New York University\\
\textsuperscript{4}CIFAR Global Scholar\\
{\tt bastings@uva.nl \{mbaroni,jase,kyunghyuncho,dkiela\}@fb.com}
}

\date{}
\begin{document}
\maketitle

\begin{abstract} \citet{lakebaroni2018scan} recently introduced the SCAN data set, which consists of simple commands paired with action sequences and is intended to test the 
strong generalization abilities
of recurrent sequence-to-sequence models. Their initial experiments suggested that such models may fail because they lack the ability to extract systematic rules.
Here, we take a closer look at SCAN and show that it does not always capture the kind of generalization that it was designed for. To mitigate this we propose a complementary dataset, which requires mapping actions back to the original commands, called NACS. We show that models that do well on SCAN do not necessarily do well on NACS, and that NACS exhibits properties more closely aligned with realistic use-cases for sequence-to-sequence models. \end{abstract}

\section{Introduction}
\label{sec:introduction}
In a recent paper, \citet{lakebaroni2018scan} (L\&B) investigate if recurrent sequence-to-sequence models can exhibit the same strong generalization that humans are capable of, by virtue of our capacity to infer the meaning of a phrase from its constituent parts (i.e.,~compositionality), providing  empirical tests for this long-standing goal \cite{fodor1988connectionism}.
Compositional generalization might be a fundamental component in making models drastically less sample-thirsty than they currently are.
L\&B introduce the SCAN data set (\S\ref{sec:scan}), meant to study such generalization to novel examples.
It consists of simple command-action pairs, in which more complex commands are composed of simpler ones (see Figure~\ref{scan-examples} for examples).

SCAN comprises several tests of generalization, namely with respect to (1) a random subset of the data (`simple'), (2) commands with action sequences longer than those seen during training (`length'), and (3) commands that compose a primitive in novel ways that was only seen in isolation during training (`primitive'). 
In the latter case, the training set would for example only include the command `jump', after which the test set includes all other commands containing `jump', e.g. `jump opposite left after walk twice'.

\input{fig/scan-examples.inp}

In this paper we take a closer look at SCAN. 
We start with the observation (\S\ref{sec:scan-prefers-simple}) that there are few target-side dependencies in the data, a consequence of SCAN being generated from a phrase-structure grammar. We show (\S\ref{sec:experiments}) that this allows simple sequence-to-sequence models (\S\ref{sec:models}) to obtain good accuracies e.g. on tasks involving a new primitive,
even without access to previous outputs. However, these simple models do not use composition in any interesting way, and their performance is therefore not a realistic indicator of their generalization capability. We hence propose NACS (\S\ref{sec:nacs}) as a more realistic alternative: SCAN with commands and actions flipped, i.e., mapping actions back to their original commands. This is harder, because different commands may map to the same action sequence, and it introduces target-side dependencies, so that previous outputs need to be remembered.

We show in particular that well-tuned attention-based models do achieve a certain degree of generalization on SCAN, and, as predicted, simpler models do better there. However, the models still struggle in the more demanding NACS setup, which we offer as a challenge for future work.

Our contributions can be summarized as follows:
\begin{enumerate}
    \item we provide an analysis of SCAN and make the important observation that it does not test for target-side dependencies, allowing too simple models to do well;
    \item we propose NACS to introduce target-side dependencies and remedy the problem;
    \item we repeat all experiments in \citet{lakebaroni2018scan} using early-stopping on validation sets created from the training data.
\end{enumerate}
\section{SCAN}
\label{sec:scan}
SCAN stands for Simplified version of the CommAI Navigation tasks \citep{mikolov2016roadmap}.
Each example in SCAN is constructed by first sampling a command ${X=(x_1, \ldots, x_T)}$ from a finite phrase-structure grammar with start symbol $C$:
\begin{flalign*}
&C \to S \textrm{ and } S \mid S \textrm{ after } S \mid S \\
&S \to V \textrm{ twice } \mid V \textrm{ thrice } \mid V\\
&V \to D_{[1]} \textrm{ opposite } D_{[2]} \mid D_{[1]} \textrm{ around } D_{[2]} \mid D \mid U\\
&D \to U \textrm{ left } \mid U \textrm{ right } \mid \textrm{ turn left } \mid \textrm{ turn right } \\
& U \to \textrm{ walk } \mid \textrm{ look } \mid \textrm{ run } \mid \textrm{ jump }
\end{flalign*}

\noindent For each command, the corresponding target action sequence ${Y=(y_1,\ldots, y_{T'})}$ then follows by applying a set of \textit{interpretation functions}, such as:
\begin{flalign*}
\llbracket \text{jump} \rrbracket =\,\,& \text{JUMP} \\ 
\llbracket u \text{ around left } \rrbracket =\,\, & \text{LTURN}\, \llbracket u \rrbracket \, \text{LTURN}\, \llbracket u\rrbracket \\
    & \text{LTURN}\, \llbracket u\rrbracket \, \text{LTURN}\, \llbracket u\rrbracket\\
\llbracket x_1 \text{ after } x_2 \rrbracket =\,\,& \llbracket x_2 \rrbracket\ \llbracket x_1\rrbracket
\end{flalign*}%
of which only the last function requires global reordering, which occurs at most once per command. See the supplementary materials for the full set.
Figure~\ref{scan-examples} shows examples of commands and their action sequences as obtained by the interpretation functions.
The commands can be decoded compositionally by a learner by discovering the interpretation functions, enabling generalization to unseen commands. The total data set is finite but large  ($20910$ unambiguous commands). 

\section{SCAN prefers simple models}
\label{sec:scan-prefers-simple}
We observe an important property of the data set generation process for SCAN: temporal dependencies of the action sequence are limited to the phrasal boundaries of each sub-phrase, which span at most 24 actions (e.g. jump around left thrice). 
Crucially, even rules that require repetition (such as `thrice') as well as global reordering, can be resolved by simple counting and without remembering previously generated outputs, due to the limited depth of the phrase-structure grammar (see e.g. \citet{rodriguez1998recurrent}).

This observation has two important implications. First, because SCAN is largely a phrase-to-phrase conversion problem, any machine learning method that aims at solving SCAN needs to have an \textit{alignment mechanism} between the source and target sequences. 
Such an alignment mechanism could work fairly accurately by simply advancing a pointer. Somewhat contrary to the observation by \citeauthor{lakebaroni2018scan} (L\&B), we therefore hypothesize that an attention mechanism~\citep{bahdanau15iclr} always helps when a neural conditional sequence model~\citep{sutskever2014sequence,cho14emnlp} is used to tackle any variant of SCAN.
Second, we speculate that any algorithm with strong long-term dependency
modeling capabilities can be \textit{detrimental} in terms of
generalization, because such an approach might inappropriately capture
spurious target-side regularities in the training data. 
We thus hypothesize that \textit{less powerful decoders generalize better} on to unseen
action combinations on SCAN when equipped with an attention mechanism.

To summarize: good performance on SCAN does not necessarily indicate the capability of a model to strongly generalize. %

SCAN favors simpler models that need not capture target-side dependencies, which might not work well on more realistic sequence-to-sequence problems, such as machine translation, where strong auto-regressive models are needed for good results~\citep{bahdanau15iclr,kaiser2016can}. 

\section{NACS: actions to commands}
\label{sec:nacs}
By simply flipping the source X and target Y of each example, we obtain a data-set that suddenly features strong target-side dependencies.  
Even when the mapping $p(Y|X)$ from the source to target is simple, the opposite $p(X|Y) \propto p(Y|X)p(X)$ is non-trivial due to the complexity of the prior $p(X)$. The inclusion of $p(X)$ naturally induces strong dependencies among the output tokens, while maintaining the original properties of SCAN that were intended to test various aspects of systematic generalization. 

NACS naturally makes the mapping that needs to be learned stochastic and multi-modal (sensitive to both commands and actions). For instance, an action sequence of the form $\llbracket x_1\rrbracket \llbracket x_2 \rrbracket$ could be mapped to either $\llbracket x_1 \text{ and } x_2 \rrbracket$ or $\llbracket x_2 \text{ after }  x_1 \rrbracket$, both of which are correct. In order for a model to decide whether to output ``and'' or ``after'', it is necessary for it to remember what has already been generated (i.e., $\llbracket x_1 \rrbracket$ or $\llbracket x_2 \rrbracket$). 

Another example is \texttt{LTURN LTURN LTURN LTURN}, which can be translated into either ``turn around left'' or two repetitions of ``turn opposite left''. 
Deciding whether to output ``and'' after the first phrase requires the model to remember whether ``around'' was generated previously.

In \S\ref{sec:experiments} we experimentally evaluate the proposed NACS task using the same 
scenarios as SCAN (simple, length and primitive).  
We observe that NACS prefers more advanced models that could capture long-term dependencies in the output (now a command sequence) better. However, we notice that even these powerful models, equipped with GRUs and attention, cannot systematically generalize to this task, as was also observed by \citet{lakebaroni2018scan}. Based on this observation, we believe that NACS (or perhaps a combination of SCAN and NACS) is better suited for evaluating any future progress in this direction.

\section{Sequence-to-sequence models}
\label{sec:models}

\begin{figure}[tb]
\centering
\input{fig/decoder.inp}
\vspace{-4mm}
\caption{\label{fig:decoder}The decoder of \citet{bahdanau15iclr}}
\vspace{-4mm}
\end{figure}

In this section, we describe the sequence-to-sequence models we use for evaluating on SCAN and its proposed sibling NACS. 

We directly model the probability of a target sequence given a source sequence $p(Y|X)$. Our encoder-decoder is modeled after \citet{cho14emnlp} and our attention-based encoder-decoder after \citet{bahdanau15iclr}. 
The attention-based decoder is a function that takes as input the previous target word embedding $\vec{e}_{y_{i-1}}$, the context vector $\vec{c}_{i}$, and the previous hidden state $\vec{s}_{i-1}$ (see also Figure~\ref{fig:decoder}):
$\vec{s}_i = f( \vec{e}_{y_{i-1}}, \vec{c}_{i}, \vec{s}_{i-1}).$

The prediction for the current time step is then made from a pre-output layer $\vec{t}_i$:
\mbox{$
\vec{t}_i = W_e \vec{e}_{y_{i-1}} + W_c \vec{c}_i + W_s \vec{s}_i.
$}
We do not apply a max-out layer and directly obtain the output by $\vec{o}_i = W_o \vec{t}_i$.
For the encoder-decoder without attention, the prediction is made directly from decoder state $\mathbf{s}_i$.
We vary the recurrent cell, experimenting with simple RNN \citep{elman1990finding}, GRU \citep{cho14emnlp}, and LSTM cells \cite{hochreiter97}. 
For conciseness we only report results with RNN and GRU cells in the main text, and LSTM results in the appendix.

In this paper, we investigate the properties of both SCAN and NACS using RNN-based sequence-to-sequence models for evaluation. We leave further investigation of alternative architectures~\citep[see, e.g.,][]{vaswani2017attention,gehring2017convolutional,chen2018best} for the future.

\section{Experiments}
\label{sec:experiments}

\begin{table*}[th]
\centering
\begin{tabular}{@{}lrrrrrrrr@{}}
\toprule
 	&\multicolumn{2}{c}{Simple} & \multicolumn{2}{c}{Length} & \multicolumn{2}{c}{Turn left} & \multicolumn{2}{c}{Jump} \\
	\cmidrule(lr){2-3} \cmidrule(lr){4-5} \cmidrule(lr){6-7} \cmidrule(lr){8-9}
\multicolumn{1}{@{}c@{}}{} & \multicolumn{1}{@{}c@{}}{SCAN} & \multicolumn{1}{@{}c@{}}{NACS} & \multicolumn{1}{@{}c@{}}{SCAN} & \multicolumn{1}{@{}c@{}}{NACS} & \multicolumn{1}{@{}c@{}}{SCAN} & \multicolumn{1}{@{}c@{}}{NACS} & \multicolumn{1}{@{}c@{}}{SCAN} & \multicolumn{1}{@{}c@{}}{NACS} \\
\midrule
\multicolumn{1}{@{}l}{GRU}& 	\nv{100.0}{0.0} & \nv{99.0}{0.1} & \nv{14.4}{0.8} & \nv{12.9}{1.2} & \nv{53.4}{11.7} & \nv{47.5}{4.7} 	&  \nv{0.0}{0.0} & \nv{0.0}{0.0} \\
\midrule
RNN \tiny{+Attn} & 
    \nv{100.0}{0.0} 	& \nv{99.8}{0.1} &
    \nv{9.6}{0.9} 	    & \nv{19.4}{0.7} & 
    \nv{81.1}{14.7} 	& \nv{44.1}{0.9} & 
    \nv{1.9}{1.2}       & \nv{0.3}{0.3}\\
RNN	\tiny{+Attn -Dep}    & 
    \nv{100.0}{0.0} 	& \nv{61.1}{0.3} & 
    \nv{11.7}{3.2} 	    & \nv{0.5}{0.2}  & 
    \nv{92.0}{5.8} 	    & \nv{18.6}{1.0} &  
    \nv{2.7}{1.7}       & \nv{0.0}{0.0}\\
GRU {\tiny +Attn} & \nv{100.0}{0.0} & \nv{99.8}{0.1}	& \nv{18.1}{1.1} & \nv{17.2}{1.9}	& \nv{59.1}{16.8} & \nv{55.9}{3.5}	& \nv{12.5}{6.6} & \nv{0.0}{0.0} \\ 
GRU {\tiny +Attn -Dep} & \nv{100.0}{0.0} & \nv{51.2}{1.2}	& \nv{17.8}{1.7} &  \nv{2.0}{1.4}	& \nv{90.8}{3.6} & \nv{16.9}{1.2}	&  \nv{0.7}{0.4} & \nv{0.0}{0.0} \\ 
\midrule
\multicolumn{1}{@{}l}{L\&B \tiny{best}}								& 99.8\hspace{18pt} &-\hspace{18pt} & 20.8\hspace{18pt} &-\hspace{18pt} & 90.3\hspace{18pt} &-\hspace{18pt} & 1.2\hspace{18pt} & -\hspace{18pt} \\
\multicolumn{1}{@{}l}{L\&B \makebox[8pt][l]{\tiny{best overall}}}	& 99.7\hspace{18pt} &-\hspace{18pt} & 13.8\hspace{18pt} &-\hspace{18pt} & 90.0\hspace{18pt} &-\hspace{18pt} & 0.1\hspace{18pt} & -\hspace{18pt} \\
\bottomrule
\end{tabular}
\caption{\label{tab:exp123combined}Test scores on the simple, length, and primitive (turn left, jump) tasks. \textit{+Attn} marks attention, \textit{-Dep} has the connections from the previous target word embedding removed ($es$ and $et$ in Figure~\ref{fig:decoder}). L\&B$_{\textrm{best}}$ is the best reported score for each task by L\&B, and L\&B$_{\textrm{best overall}}$ is the score for their best-scoring model all tasks considered.}
\end{table*}

\subsection{Settings}

Our models are implemented in PyTorch and trained using mini-batch SGD with an initial learning rate of $0.2$, decayed by $0.96$ each epoch. We use a batch size of $32$, $256$ hidden units ($64$ for embeddings), and a dropout rate of $0.2$. 
We test on all SCAN/NACS tasks\footnote{\url{github.com/facebookresearch/NACS}},
as well as on the Fr-En Machine Translation (MT) task that L\&B used.
The reported results are averaged over three runs for each experiment. Models \textit{with} attention are marked as such with \textit{+Attn}, e.g. `GRU {+Attn}'. 

\paragraph{Validation Set.} 
L\&B split each SCAN subtask into a training set (80\%) and a test set (20\%). They train for a fixed number of updates (100k) and evaluate on the test set. 
Because any training run without early stopping may have missed the optimal solution \citep{caruana2001overfitting}, we believe their results may not reflect the reality as closely as they could.
We thus augment each of the SCAN variants with a validation set that follows the training distribution but contains examples that are not contained in the corresponding training set. This allows us to incorporate early stopping in our experiments so that they are better benchmarks for evaluating future progress. For each experiment we remove 10\% of the training examples to be used as a validation set.

\paragraph{Accuracy.} Following L\&B we measure performance according to sequence-level accuracy, i.e., whether the generated sequence entirely matches the reference. This metric is also used for early stopping. For NACS, an output (command) is considered correct if its interpretation (`back-mapping') produces the input action sequence.

\paragraph{Ablations.} To validate our analysis, we remove the connections from the previous target word embedding $\mathbf{e}_{y_{i-1}}$ to the decoder state and the pre-output layer ($es$ and $et$ in Figure~\ref{fig:decoder}), so that the current prediction is not informed by previous outputs. 
If our analysis in \S\ref{sec:scan-prefers-simple} is correct, then these simpler models should still be able to make the correct predictions on SCAN, but not on NACS.

\subsection{Results and Analysis}

Results on the three SCAN and NACS tasks are listed in Table~\ref{tab:exp123combined}. The full results including models with LSTM cells and MT experiments may be found in the supplementary materials. We will now discuss our observations.

\paragraph{SCAN is not enough.} Table~\ref{tab:exp123combined} shows that all model variants perform (near) perfectly on the SCAN simple task. While this is impressive, results for the severed models (+Attn -Dep) on the simple task for NACS show that it is possible to have a perfect accuracy on SCAN, while at the same time failing to do well on NACS.\footnote{We made similar observations using LSTM cells, as we show in the appendix.} Crucially, a (near) perfect score on SCAN does not imply strong generalization. A model can exploit the determinism and lack of target-side dependencies of SCAN by developing a simple translation strategy such as advancing a pointer and translating word by word, and the use of such a simple strategy is not revealed by SCAN.

\paragraph{NACS is harder.}
NACS is a harder problem to solve compared to SCAN, as evidenced by consistently lower accuracies in Table~\ref{tab:exp123combined} for all tasks. The discrepancy between SCAN and NACS performance is the most extreme when we look at the primitive tasks (turn left and jump). For turn left, the severed models (+Attn -Dep) obtain the highest scores on SCAN, but are the worst on NACS. 

The `turn left' task benefits from \texttt{TURNL} occurring on the target-side in other contexts during training, which is not the case for `jump'.\footnote{See \citet{lakebaroni2018scan} for a discussion.} Since there is no evidence in the training data that `jump' is a verb, Table~\ref{tab:expjumpaddcombined} shows results where additional (composed) `jump' commands were added for training.  
We see that performance quickly goes up when adding more commands.\footnote{L\&B performed this experiment without attention, which we show has a large positive impact.}  
Again here the simpler models (+Attn -Dep) perform better.

\begin{table*}[t!]
\centering
\begin{tabular}{@{}llrrrrrr@{}}
\toprule
& & 
\multicolumn{1}{c}{1} & \multicolumn{1}{c}{2} & \multicolumn{1}{c}{4} & \multicolumn{1}{c}{8} & \multicolumn{1}{c}{16} & \multicolumn{1}{c}{32} \\
\midrule
RNN \tiny{+Attn}     & SCAN 	& \nv{35.0}{2.8}	& \nv{48.6}{8.1}	& \nv{77.6}{2.6}	& \nv{89.2}{3.8}	& \nv{98.7}{1.3}	& \nv{99.8}{0.1} \\ 
RNN	\tiny{+Attn -Dep} & SCAN	& \nv{29.5}{10.5}	& \nv{53.3}{10.2}	& \nv{82.4}{4.7}	& \nv{98.8}{0.8}	& \nv{99.8}{0.1}	& \nv{100.0}{0.0} \\
GRU {\tiny +Attn} 				& SCAN & 
\nv{58.2}{12.0}	& \nv{67.8}{3.4}	& \nv{80.3}{7.0}	& \nv{88.0}{6.0}	& \nv{98.3}{1.8}	&  \nv{99.6}{0.2} \\
GRU {\tiny +Attn -Dep} & SCAN & 
\nv{70.9}{11.5}	& \nv{61.3}{13.5}	& \nv{83.5}{6.1}	& \nv{99.0}{0.4}	& \nv{99.7}{0.2}	& \nv{100.0}{0.0} \\
\midrule
RNN \tiny{+Attn} & NACS	& \nv{2.8}{0.8}		& \nv{9.3}{7.3}	&\nv{24.7}{4.2}	&\nv{43.7}{4.4}	&\nv{57.1}{5.2}	&\nv{69.1}{2.1}\\ 
RNN \tiny{+Attn -Dep} & NACS	& \nv{0.4}{0.1}		& \nv{0.9}{0.2}	& \nv{2.4}{0.3}	& \nv{3.9}{0.3}	& \nv{9.3}{0.3}	&\nv{15.9}{1.4}\\
GRU {\tiny +Attn}				& NACS			&
\nv{5.5}{1.8}	& \nv{9.2}{2.8}	& \nv{11.0}{1.5}& \nv{21.9}{2.4}	& \nv{23.5}{0.6}& \nv{42.0}{1.5}\\
GRU {\tiny +Attn -Dep} & NACS &  
\nv{0.1}{0.1} & \nv{0.6}{0.2} & \nv{2.0}{0.2} & \nv{3.2}{0.2} & \nv{5.8}{1.1}	& \nv{10.9}{0.8} \\
\midrule
L\&B 			& SCAN &  
0.1\hspace{18pt} & 0.1\hspace{18pt} & 4.1\hspace{18pt} & 15.3\hspace{18pt} & 70.2\hspace{18pt} & 89.9\hspace{18pt} \\
\bottomrule
\end{tabular}
\caption{\label{tab:expjumpaddcombined}Test scores on the `jump' task with additional commands. 
+Attn marks attention, -Dep has the $es$ and $et$ connections removed (Figure~\ref{fig:decoder}).
The test set contains all jump commands except the 32 used for training. 
Columns indicate how many commands with `jump' were added to the training set, such as `jump around left thrice.'
}
\vspace{-4mm}
\end{table*}
\begin{table}[t!]
\centering
\begin{tabular}{lrr}
\toprule
{} &\multicolumn{1}{c}{En-Fr} & \multicolumn{1}{c}{Fr-En} \\ \midrule
GRU	\tiny{+Attn} & \nv{32.1}{0.3} 	& \nv{37.5}{0.6} \\
GRU	\tiny{+Attn -Dep}	& \nv{30.2}{0.3} 	& \nv{35.9}{0.3} \\
\bottomrule
\end{tabular}
\caption{\label{tab:mt}Results (BLEU) on the Machine Translation experiment for both directions using a GRU. See appendix for results using SRN and LSTM cells.}
\end{table}

\paragraph{Machine Translation.} We repeat L\&Bs English-French MT experiment for both directions. 
Table~\ref{tab:mt} shows that models that perform well on NACS also perform well here, with the GRU outperforming the other cells (see appendix for other cell types).
In a setting similar to the jump task, the sentence pair `I am daxy' (`je suis daxiste') was added to the training set.
The goal is now to test if eight novel sentences that contain `daxy' are correctly translated. 

In our setting with mini-batching and early-stopping, the GRU gets 70.8\% (En-Fr) and 54.2\% (Fr-En) of the daxy-sentences right, which is surprisingly good compared to L\&B (12.5\%).

\paragraph{Other observations.} As expected, Table~\ref{tab:exp123combined} shows that attention always helps. Generalizing to longer sequences is generally hard, and this remains an open problem.

\section{Related Work}
Ever since \citet{fodor1988connectionism} conjectured that neural networks are unable to show strong generalization, many attempts were made to show that the opposite is true, leading to inconclusive evidence.
For example, \citet{phillips1998feedforward} found that feed-forward nets and RNNs do not always generalize to novel 2-tuples 
on an auto-association task, while \citet{wong2007generalisation} and \citet{brakel2009strong} found that RNNs can show systematic behavior in a language modeling task.

In the context of analyzing RNNs, \citet{rodriguez1998recurrent} found that simple RNNs can develop a symbol-sensitive counting strategy for accepting a simple (palindrome) context-free language. 
\citet{weiss2018practical} show that LSTMs and simple RNNs with ReLU-activation can learn to count unboundedly, in contrast to GRUs.

\citet{linzen2016lstmsyntax} probed the sensitivity of LSTMs to hierarchical structure (not necessarily in novel constructions). 
Instead of a binary choice, with SCAN a sequence-to-sequence model productively generates an output string.

\citet{liska2018haystack} found that a small number of identical RNNs trained with different initializations show compositional behavior on a function composition task, suggesting that more specific architectures may not be necessary.

Finally, \citet{lakebaroni2018scan} introduced the SCAN data set to study systematic compositionality in recurrent sequence-to-sequence models, including gating mechanisms and attention. This work is a direct response to that and aims to facilitate future progress by showing that SCAN does not necessarily test for strong generalization.

\section{Conclusion}
\label{sec:conclusion}
In the quest for strong generalization, benchmarks measuring progress are an important component. The existing SCAN benchmark allows too simple models to shine, without the need for compositional generalization.
We proposed NACS to remedy this. NACS still requires systematicity, while introducing 
stochasticity and strong dependencies on the target side.
We argue that a good benchmark needs at least those properties, in order not to fall prey to trivial solutions, which do not work on more realistic use-cases for sequence-to-sequence models such as machine translation. 

\newpage

\section*{Acknowledgments}
We would like to thank Brenden Lake and Marc'Aurelio Ranzato for useful discussions and feedback.

\bibliography{bib}
\bibliographystyle{acl_natbib_nourl}

\appendix
\newpage
\begin{figure*}
\begin{center}
\huge Supplementary Materials
\end{center}
\end{figure*}

\input{suppl/interpretation.inp}

\newpage

\begin{table*}[p!]
\centering
\begin{tabular}{lrrrr}
\toprule
{}    & \multicolumn{1}{c}{Simple} & \multicolumn{1}{c}{Length} & \multicolumn{1}{c}{Turn left} & \multicolumn{1}{c}{Jump} \\
\midrule
RNN & \nv{75.6}{5.4} 	&  \nv{0.2}{0.0} 	& \nv{26.7}{12.8} 	&  \nv{0.0}{0.0} \\
GRU	& \nv{100.0}{0.0} 	& \nv{14.4}{0.8} 	& \nv{53.4}{11.7} 	&  \nv{0.0}{0.0} \\
LSTM& \nv{99.8}{0.1} 	& \nv{10.1}{2.0} 	& \nv{56.5}{0.8} 	&  \nv{0.1}{0.0} \\
\midrule
RNN \tiny{+Attn} & \nv{100.0}{0.0} 	&  \nv{9.6}{0.9} 	& \nv{81.1}{14.7} 	&  \nv{1.9}{1.2} \\
RNN	\tiny{+Attn-Dep} & \nv{100.0}{0.0} 	& \nv{11.7}{3.2} 	& \nv{92.0}{5.8} 	&  \nv{2.7}{1.7} \\ 
GRU	\tiny{+Attn} & \nv{100.0}{0.0} 	& \nv{18.1}{1.1} 	& \nv{59.1}{16.8} 	& \nv{12.5}{6.6} \\ 
GRU \tiny{+Attn-Dep}		& \nv{100.0}{0.0} 	& \nv{17.8}{1.7} 	& \nv{90.8}{3.6} 	&  \nv{0.7}{0.4} \\ 
LSTM \tiny{+Attn} & \nv{100.0}{0.0} 	& \nv{15.6}{1.6} 	& \nv{83.8}{16.8} 	&  \nv{9.7}{2.9} \\
LSTM \tiny{+Attn-Dep}	& \nv{100.0}{0.0} 	& \nv{12.5}{1.3} 	& \nv{57.6}{3.8} 	&  \nv{0.8}{0.5} \\ 
\midrule
L\&B \makebox[8pt][l]{\tiny{best}} & 99.8\hspace{18pt} & 20.8\hspace{18pt} & 90.3\hspace{18pt} & 1.2\hspace{18pt} \\
L\&B \makebox[8pt][l]{\tiny{best overall}} & 99.7\hspace{18pt} & 13.8\hspace{18pt} & 90.0\hspace{18pt} & 0.1\hspace{18pt} \\
\bottomrule
\end{tabular}
\caption{\label{tab:suppl-exp123}SCAN test scores on the simple, length, and primitive (turn left and jump) tasks. For `+Attn-Dep' models we removed the connections from the previous target word embedding to the decoder state and the pre-output layer.}
\end{table*}

\begin{table*}[p]
\centering
\begin{tabular}{lrrrr}
\toprule
{}    & \multicolumn{1}{c}{Simple} & \multicolumn{1}{c}{Length} & \multicolumn{1}{c}{Turn left} & \multicolumn{1}{c}{Jump} \\
\midrule
RNN	& 	\nv{26.9}{0.2} &  \nv{0.2}{0.1} & \nv{26.4}{12.0} & \nv{0.0}{0.0}\\
GRU	& 	\nv{99.0}{0.1} & \nv{12.9}{1.2} & \nv{47.5}{4.7}  & \nv{0.0}{0.0}\\
LSTM& 	\nv{99.1}{0.1} & \nv{10.9}{1.3} & \nv{42.9}{2.9}  & \nv{0.0}{0.0}\\
\midrule
RNN \tiny{+Attn} & \nv{99.8}{0.1} & \nv{19.4}{0.7} & \nv{44.1}{0.9} & \nv{0.3}{0.3}\\
RNN	\tiny{+Attn-Dep} & \nv{61.1}{0.3} &  \nv{0.5}{0.2} & \nv{18.6}{1.0} & \nv{0.0}{0.0}\\ 
GRU	\tiny{+Attn} & \nv{99.8}{0.1} & \nv{17.2}{1.9} & \nv{55.9}{3.5} & \nv{0.0}{0.0}\\ 
GRU	\tiny{+Attn-Dep}	& \nv{51.2}{1.2} &  \nv{2.0}{1.4} & \nv{16.9}{1.2} & \nv{0.0}{0.0}\\ 
LSTM \tiny{+Attn} & \nv{99.1}{0.2} & \nv{17.1}{2.0} & \nv{48.3}{1.7} & \nv{0.0}{0.0}\\ 
LSTM \tiny{+Attn-Dep}	& \nv{38.9}{0.9} &  \nv{1.0}{0.5} & \nv{17.2}{1.2} & \nv{0.0}{0.0}\\ 
\bottomrule
\end{tabular}
\caption{\label{tab:suppl-exp123-nacs}NACS test scores on the simple, length, and primitive (turn left and jump) tasks. For `+Attn-Dep' models we removed the connections from the previous target word embedding to the decoder state and the pre-output layer.}
\end{table*}

\begin{table*}[p]
\setlength{\tabcolsep}{5pt}
\centering
\begin{tabular}{@{}lrrrrrrr@{}}
\toprule
{} & \multicolumn{1}{c}{0} & \multicolumn{1}{c}{1} & \multicolumn{1}{c}{2} & \multicolumn{1}{c}{4} & \multicolumn{1}{c}{8} & \multicolumn{1}{c}{16} & \multicolumn{1}{c}{32} \\ \midrule
RNN & \nv{0.0}{0.0}	& \nv{0.0}{0.0}	& \nv{0.0}{0.0}	& \nv{0.1}{0.0}	& \nv{0.1}{0.1}	& \nv{0.5}{0.3}	& \nv{1.4}{0.3}\\
GRU & \nv{0.1}{0.0}	& \nv{0.2}{0.1}	& \nv{0.6}{0.2}	& \nv{2.5}{1.1}	& \nv{3.3}{0.9}	& \nv{13.1}{2.4}	& \nv{42.4}{2.5}\\
LSTM& \nv{0.1}{0.0}	& \nv{0.3}{0.2}	& \nv{1.3}{0.2}	& \nv{3.8}{1.8}	& \nv{2.5}{1.1}	& \nv{6.5}{2.7}	& \nv{21.3}{1.4}\\
\midrule
RNN \tiny{+Attn} & \nv{3.5}{3.0}	& \nv{35.0}{2.8}	& \nv{48.6}{8.1}	& \nv{77.6}{2.6}	& \nv{89.2}{3.8}	& \nv{98.7}{1.3}	& \nv{99.8}{0.1} \\ 
RNN	\tiny{+Attn-Dep} 	& \nv{2.7}{1.7}	& \nv{29.5}{10.5}	& \nv{53.3}{10.2}	& \nv{82.4}{4.7}	& \nv{98.8}{0.8}	& \nv{99.8}{0.1}	& \nv{100.0}{0.0} \\
GRU	\tiny{+Attn} & \nv{12.5}{6.6}	& \nv{58.2}{12.0}	& \nv{67.8}{3.4}	& \nv{80.3}{7.0}	& \nv{88.0}{6.0}	& \nv{98.3}{1.8}	&  \nv{99.6}{0.2} \\
GRU	\tiny{+Attn-Dep} 	& \nv{0.7}{0.4}	& \nv{70.9}{11.5}	& \nv{61.3}{13.5}	& \nv{83.5}{6.1}	& \nv{99.0}{0.4}	& \nv{99.7}{0.2}	& \nv{100.0}{0.0} \\
LSTM \tiny{+Attn} & \nv{7.8}{0.9}	& \nv{40.2}{9.3}	& \nv{37.7}{10.7}	& \nv{50.3}{13.9}	& \nv{62.2}{7.7}	& \nv{94.0}{2.7}	& \nv{98.6}{1.0} 
\\
LSTM \tiny{+Attn-Dep} & \nv{0.8}{0.6}	& \nv{39.0}{6.5}	& \nv{43.6}{17.6}	& \nv{66.0}{1.6}	& \nv{86.1}{2.3}	& \nv{98.7}{1.6}	& \nv{99.8}{0.2} \\

\midrule
L\&B 			& 0.1\hspace{18pt} & 0.1\hspace{18pt} & 0.1\hspace{18pt} &  4.1\hspace{18pt} & 15.3\hspace{18pt} & 70.2\hspace{18pt} & 89.9\hspace{18pt} \\
\bottomrule
\end{tabular}
\caption{\label{tab:suppl-jump-add}SCAN test scores for jump with additional composed commands.}
\end{table*}

\begin{table*}[p]
\setlength{\tabcolsep}{5pt}
\centering
\begin{tabular}{@{}lrrrrrrr@{}}
\toprule
{}  & \multicolumn{1}{c}{0} & \multicolumn{1}{c}{1} & \multicolumn{1}{c}{2} & \multicolumn{1}{c}{4} & \multicolumn{1}{c}{8} & \multicolumn{1}{c}{16} & \multicolumn{1}{c}{32} \\ \midrule
RNN & \nv{0.0}{0.0}	& \nv{0.1}{0.0}	& \nv{0.1}{0.1}	& \nv{0.2}{0.0}	& \nv{0.7}{0.2}	& \nv{0.4}{0.0}	& \nv{0.8}{0.1}\\
GRU & \nv{0.0}{0.0}	& \nv{0.3}{0.2}	& \nv{0.4}{0.1}	& \nv{0.3}{0.2}	& \nv{1.0}{0.4}	& \nv{5.8}{0.1}	&\nv{20.8}{2.2}\\
LSTM& \nv{0.0}{0.0}	& \nv{0.6}{0.4}	& \nv{0.5}{0.3}	& \nv{0.7}{0.0}	& \nv{1.0}{0.3}	& \nv{3.7}{0.4}	&\nv{11.4}{1.2}\\
\midrule
RNN \tiny{+Attn} &\nv{0.3}{0.3}	& \nv{2.8}{0.8}		& \nv{9.3}{7.3}	&\nv{24.7}{4.2}	&\nv{43.7}{4.4}	&\nv{57.1}{5.2}	&\nv{69.1}{2.1}\\ 
RNN \tiny{+Attn-Dep} & \nv{0.0}{0.0}	& \nv{0.4}{0.1}		& \nv{0.9}{0.2}	& \nv{2.4}{0.3}	& \nv{3.9}{0.3}	& \nv{9.3}{0.3}	&\nv{15.9}{1.4}\\
GRU	\tiny{+Attn} &\nv{0.0}{0.0}	& \nv{5.5}{1.8}		& \nv{9.2}{2.8}	&\nv{11.0}{1.5}	&\nv{21.9}{2.4}	&\nv{23.5}{0.6}	&\nv{42.0}{1.5}\\
GRU	\tiny{+Attn-Dep} 	&\nv{0.0}{0.0}	& \nv{0.1}{0.1}		& \nv{0.6}{0.2}	& \nv{2.0}{0.2}	& \nv{3.2}{0.2}	& \nv{5.8}{1.1}	&\nv{10.9}{0.8}\\
LSTM \tiny{+Attn} & \nv{0.0}{0.0}	& \nv{2.1}{0.2}		& \nv{3.7}{0.9}	& \nv{6.6}{0.5}	&\nv{12.5}{2.5}	&\nv{21.8}{2.6}	&\nv{34.2}{1.7}\\
LSTM \tiny{+Attn-Dep} &\nv{0.0}{0.0}	& \nv{0.4}{0.2}		& \nv{0.9}{0.1}	& \nv{1.5}{0.2}	& \nv{1.9}{0.3}	& \nv{3.2}{0.6}	& \nv{7.4}{0.9}\\
\bottomrule
\end{tabular}
\caption{\label{tab:suppl-jump-add-nacs}NACS test scores for jump with additional composed commands.}
\end{table*}

\begin{table*}[p]
\centering
\begin{tabular}{lrr}
\toprule
{} &\multicolumn{1}{c}{En-Fr} & \multicolumn{1}{c}{Fr-En} \\ \midrule
RNN \tiny{+Attn} & \nv{29.1}{0.4} 	& \nv{34.9}{0.8} \\
RNN	\tiny{+Attn-Dep}	& \nv{27.5}{0.7} 	& \nv{32.9}{0.8} \\
GRU	\tiny{+Attn} & \nv{32.1}{0.3} 	& \nv{37.5}{0.6} \\
GRU	\tiny{+Attn-Dep}	& \nv{30.2}{0.3} 	& \nv{35.9}{0.3} \\
LSTM \tiny{+Attn} 	 & \nv{31.5}{0.2} 	& \nv{36.9}{1.1} \\
LSTM \tiny{+Attn-Dep}& \nv{28.7}{0.2} 	& \nv{34.0}{0.1} \\
\bottomrule
\end{tabular}
\caption{\label{tab:suppl-mt}Results (BLEU) on the Machine Translation experiment for both directions.}
\end{table*}

\begin{table*}[p]
\centering
\begin{tabular}{lrr} \toprule
{} &\multicolumn{1}{c}{En-Fr} & \multicolumn{1}{c}{Fr-En} \\ \midrule
RNN \tiny{+Attn} & \nv{79.2}{15.6 }	& \nv{41.7}{ 5.9}\\
RNN	\tiny{+Attn-Dep} 	& \nv{66.7}{ 5.9 }	& \nv{41.7}{ 5.9}\\
GRU	\tiny{+Attn} & \nv{70.8}{11.8 }	& \nv{54.2}{ 5.9}\\
GRU	\tiny{+Attn-Dep} 	& \nv{58.3}{ 5.9 }	& \nv{45.8}{11.8}\\
LSTM \tiny{+Attn} & \nv{75.0}{10.2 }	& \nv{41.7}{15.6}\\
LSTM \tiny{+Attn-Dep} & \nv{50.0}{10.2 }	& \nv{41.7}{ 5.9}\\
\bottomrule
\end{tabular}
\caption{\label{tab:suppl-mt-daxy}Machine Translation: accuracy on eight novel sentences containing `daxy' (`daxiste').}
\end{table*}

\begin{table*}[p]
\centering
\begin{tabular}{lrr} \toprule
{} &\multicolumn{1}{c}{En-Fr} & \multicolumn{1}{c}{Fr-En} \\ \midrule
RNN \tiny{+Attn} & \nv{66.7}{ 5.9}	& \nv{20.8}{ 5.9}\\
RNN	\tiny{+Attn-Dep}	& \nv{66.7}{ 5.9}	& \nv{29.2}{15.6}\\ 
GRU	\tiny{+Attn} & \nv{62.5}{ 0.0}	& \nv{33.3}{ 5.9}\\
GRU	\tiny{+Attn-Dep}	& \nv{66.7}{ 5.9}	& \nv{25.0}{ 20.4}	\\ 
LSTM \tiny{+Attn} & \nv{66.7}{ 5.9}	& \nv{25.0}{10.2}\\
LSTM \tiny{+Attn-Dep} & \nv{62.5}{ 0.0}	& \nv{25.0}{17.7}\\ 
\bottomrule
\end{tabular}
\caption{\label{tab:suppl-mt-tired}Machine Translation: accuracy on eight novel sentences containing `tired' (`fatigu\'e').}
\end{table*}

\end{document}